\documentclass[10pt,twocolumn,letterpaper]{article}

\usepackage[pagenumbers]{cvpr} % To force page numbers, e.g. for an arXiv version

\usepackage{xspace}
\usepackage[dvipsnames]{xcolor}

\newcommand{\name}{KYN\xspace}

\usepackage{acro}
\DeclareAcronym{vl}{
  short = VL ,
  long = vision-language
}
\DeclareAcronym{kyn}{
  short = KYN ,
  long = Know Your Neighbors
}

\DeclareAcronym{vlm}{
    short = VLM ,
    long = vision-language modulation
}

\DeclareAcronym{vlsa}{
    short = VLSA ,
    long = vision-language spatial attention
}

\definecolor{cvprblue}{rgb}{0.21,0.49,0.74}
\usepackage[pagebackref,breaklinks,colorlinks,citecolor=cvprblue]{hyperref}
\usepackage{array} % For better handling of columns in tables
\usepackage{colortbl}
\usepackage{booktabs}
\usepackage{multirow}
\usepackage{amsmath}
\usepackage{graphicx}
\usepackage{caption}
\usepackage{subcaption}
\usepackage{adjustbox}
\usepackage{rotating}
\DeclareMathOperator*{\argmax}{arg\,max}
\usepackage{xcolor}
\usepackage[accsupp]{axessibility} % Improves PDF readability for those with visual impairments.
\definecolor{teasergreen}{rgb}{0.13, 0.55, 0.13}
\colorlet{colorFst}{Green!25}       % first
\colorlet{colorSnd}{SpringGreen!45} % second
\colorlet{colorTrd}{Yellow!30}      % third
\newcommand{\fs}{\cellcolor{colorFst}\bf}   % first
\newcommand{\nd}{\cellcolor{colorSnd}}      % second
\newcommand{\rd}{\cellcolor{colorTrd}}      % third

\def\paperID{2887} % *** Enter the Paper ID here
\def\confName{CVPR}
\def\confYear{2024}

\title{Know Your Neighbors: Improving Single-View Reconstruction \\ via Spatial Vision-Language Reasoning}

\author{Rui Li\textsuperscript{1} \quad Tobias Fischer\textsuperscript{1} \quad Mattia Segu\textsuperscript{1}\quad Marc Pollefeys\textsuperscript{1} \\ Luc Van Gool\textsuperscript{1} \quad Federico Tombari\textsuperscript{2, 3}\\
\textsuperscript{1}ETH Zürich \quad \textsuperscript{2}Google \quad 
\textsuperscript{3}Technical University of Munich\\
}

\begin{document}
\maketitle

\begin{abstract}
% problem
Recovering the 3D scene geometry from a single view is a fundamental yet ill-posed problem in computer vision. 
% technical gap
While classical depth estimation methods infer only a 2.5D scene representation limited to the image plane, recent approaches based on radiance fields reconstruct a full 3D representation. However, these methods still struggle with occluded regions since inferring geometry without visual observation requires (i) semantic knowledge of the surroundings, and (ii) reasoning about spatial context. 
% our contributions
We propose \name, a novel method for single-view scene reconstruction that reasons about semantic and spatial context to predict each point's density. 
We introduce a vision-language modulation module to enrich point features with fine-grained semantic information. 
We aggregate point representations across the scene through a language-guided spatial attention mechanism to yield per-point density predictions aware of the 3D semantic context.
% results
We show that \name improves 3D shape recovery compared to predicting density for each 3D point in isolation. We achieve state-of-the-art results in scene and object reconstruction on KITTI-360, and show improved zero-shot generalization compared to prior work.
% code
{Project page: \href{https://ruili3.github.io/kyn}{https://ruili3.github.io/kyn}}.

\end{abstract}    
\section{Introduction}
\label{sec:intro}

\begin{figure}[tp]
\centering
% \vspace{-12pt}
\subfloat{
    \begin{minipage}[c]{0.03\linewidth}
        \centering
        % \scriptsize
        \rotatebox{90}{\makebox{Input}}
    \end{minipage}
    \begin{minipage}[c]{0.9\linewidth}
        \centering
        \includegraphics[width=1\linewidth]{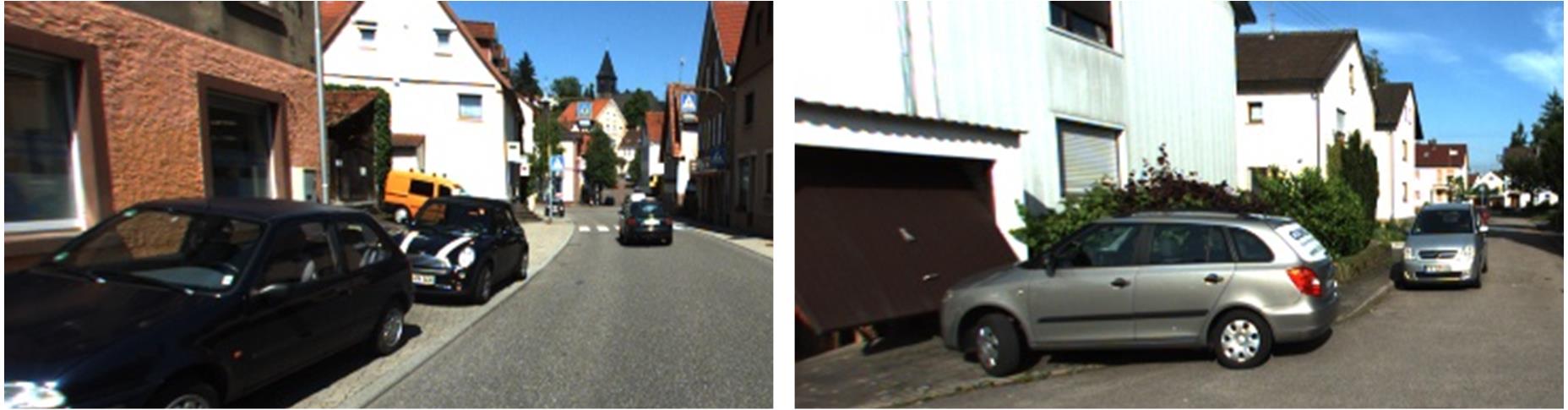}
    \end{minipage}
} \vspace{0.5pt}\\
\subfloat{
    \begin{minipage}[c]{0.03\linewidth}
        \centering
        % \scriptsize
        \rotatebox{90}{\makebox{BTS \cite{wimbauer2023behind}}}
    \end{minipage}
    \begin{minipage}[c]{0.9\linewidth}
        \centering
        \includegraphics[width=1\linewidth]{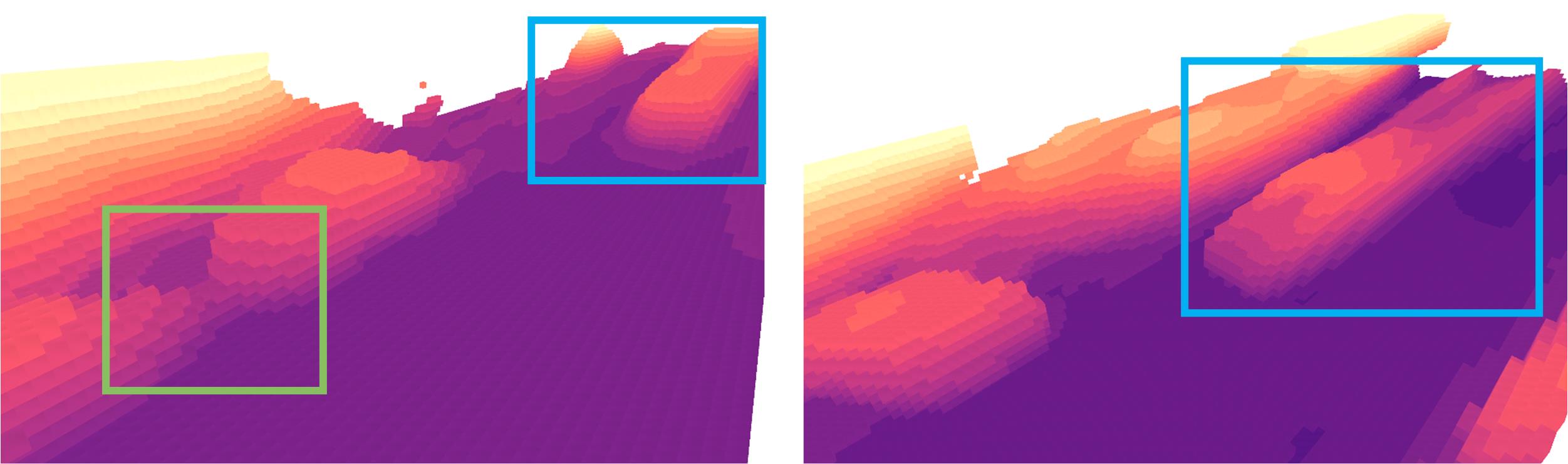}
    \end{minipage}
} \vspace{0.5pt}\\
\subfloat{
    \begin{minipage}[c]{0.03\linewidth}
        \centering
        % \scriptsize
        \rotatebox{90}{\makebox{\textbf{Ours}}}
    \end{minipage}
    \begin{minipage}[c]{0.9\linewidth}
        \centering
        \includegraphics[width=1\linewidth]{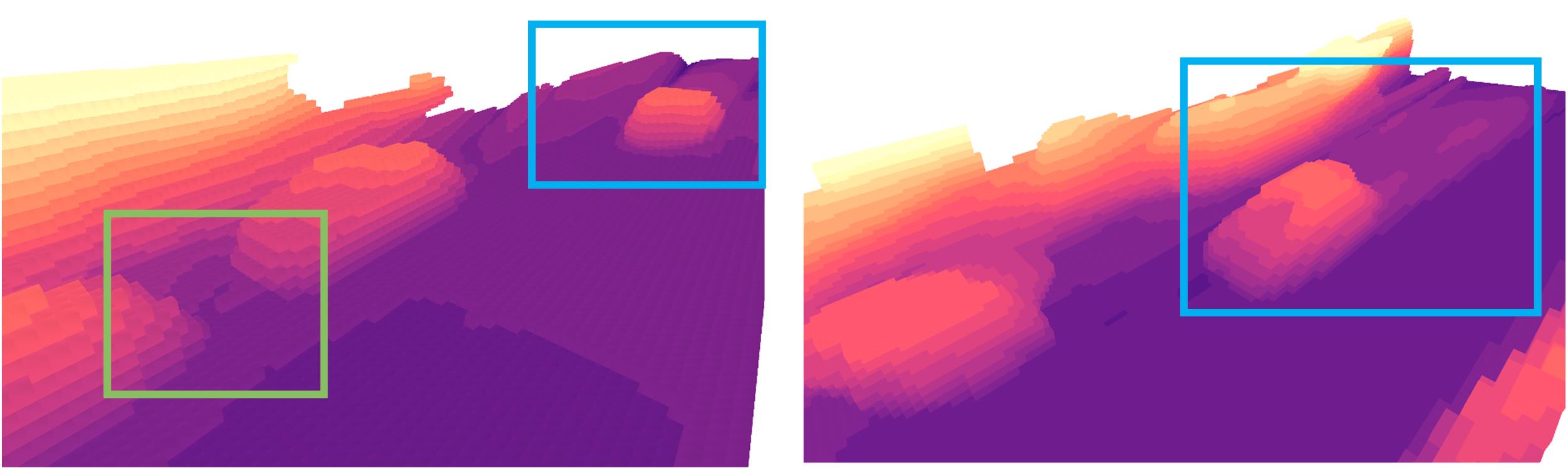}
    \end{minipage}
} \vspace{0.5pt}\\
\vspace{-5pt}
\caption{\textbf{Single-view scene reconstruction results.} We present the predicted 3D occupancy grids given a single input image. The camera is at the bottom left and points to the top right along the $z$-aixs. Previous methods like BTS~\cite{wimbauer2023behind} struggle to recover accurate object shapes (\textcolor{teasergreen}{green} box) and exhibit trailing effects in unobserved areas (\textcolor{blue}{blue} box). In contrast, \name recovers more accurate boundaries and mitigates the trailing effects prevalent in prior art.}
% \vspace{-3pt}
\label{fig:teaser}
\end{figure}

Humans have the extraordinary ability to estimate the geometry of a 3D scene from a single image, often including its occluded parts. It enables us to reason about where dynamic actors in the scene might move, and how to best navigate ourselves to avoid a collision. Hence, estimating the 3D scene geometry from a single input view is a long-standing challenge in computer vision, fundamental to autonomous navigation \cite{geiger2012we} and virtual reality applications \cite{liang2023movideo}. 
Since the problem is highly ill-posed due to scale ambiguity, occlusions, and perspective distortion, it has traditionally been cast as a 2.5D problem \cite{yin2023metric3d, li2023learning, cheng2024coatrsnet}, focusing on areas visible in the image plane and neglecting the non-visible parts.

\par
Recently, approaches based on neural radiance fields~\cite{mildenhall2020nerf} have shown great potential in inferring the true 3D scene representation from a single~\cite{yu2021pixelnerf, wimbauer2023behind} or multiple views~\cite{wang2021ibrnet}. For instance, Wimbauer~\etal~\cite{wimbauer2023behind} introduce BTS, a method that estimates a 3D density field from a single view at inference while being supervised only by photometric consistency given multiple posed views at training time.
Intuitively, given only a single image at inference, the model must rely on semantic knowledge from the neighboring 3D structure to predict the density of occluded points. 
However, existing approaches lack explicit semantic modeling and, by modeling density prediction independently for each point, are unaware of the semantic 3D context of the point's surroundings.
This results in the clear limitations which we illustrate in \cref{fig:teaser}. Specifically, prior work~\cite{wimbauer2023behind} struggles with accurate shape recovery (\textcolor{teasergreen}{green}) and further exhibits trailing effects (\textcolor{blue}{blue}) in the absence of visual observation.
We argue that, when considering a single point in 3D, its density highly depends on the semantic scene context, \eg if there is an intersection, a parking lot, or a sidewalk visible in its proximity. This becomes more critical as we move further from the camera origin since the degree of visual coverage decreases with distance, and reconstructing the increasingly unobserved scene parts requires context from the neighboring points.

To this end, we present \ac{kyn}, a novel approach for single-view scene reconstruction that predicts density for each 3D point in a scene by reasoning about its neighboring semantic and spatial context. 
We introduce two key innovations.
We develop a \ac{vl} modulation scheme that endows the representation of each 3D point in space with fine-grained semantic information. To leverage this information, we further introduce a \ac{vl} spatial attention mechanism that utilizes language guidance to aggregate the visual semantic point representations across the scene and predict the density of each individual point as a function of the neighboring semantic context. 

We show that, by injecting semantic knowledge and reasoning about spatial context, our method overcomes the limitations that prior art exhibits in unobserved areas, producing more plausible 3D shapes and mitigating their trailing effects.
We summarize our contributions as follows:
% \vspace{-0.2cm}
\begin{itemize}
\item We propose \ac{kyn}, the first single-view scene reconstruction method that reasons about semantic and spatial context to predict each point's density.
\item We introduce a \ac{vl} modulation module to enrich point features with fine-grained semantic information.
\item We propose a \ac{vl} spatial attention mechanism that aggregates point representations across the scene to yield per-point density predictions aware of the neighboring 3D semantic context.

\end{itemize}
Our experiments on the KITTI-360 dataset~\cite{liao2022kitti} show that \ac{kyn} achieves state-of-the-art scene and object reconstructions. Furthermore, we demonstrate that \ac{kyn} exhibits better zero-shot generalization on the DDAD dataset~\cite{guizilini20203d} compared to the prior art.
\section{Related Work}
\label{sec:related_work}
\noindent \textbf{Monocular depth estimation.} 
Estimating depth from a single view has been extensively studied over the last decade~ \cite{yin2019enforcing, yin2023metric3d, yin2022towards, yin2021learning, liu2023va, liu2023single}, both in a supervised and a self-supervised manner. Supervised methods directly minimize the loss between the predicted and ground truth depths~\cite{eigen2014depth,bhat2021adabins}. For these, varying output representations \cite{fu2018deep,bhat2021adabins,bhat2022localbins}, network architectures \cite{bhat2021adabins,yuan2022new, ranftl2021vision, lee2022edgeconv}, and loss functions \cite{bhat2021adabins,xian2020structure,teed2018deepv2d} have been proposed. Recent methods explore training unified depth models on large datasets, tackling challenges like varying camera intrinsics~\cite{ranftl2020towards,yin2021learning, yin2023metric3d, facil2019cam,guizilini2023towards} and dataset bias \cite{bhat2023zoedepth}. 
Self-supervised methods cast the problem as a view synthesis task and learn depth via photometric consistency on image pairs. Existing works have investigated how to handle dynamic objects~\cite{li2023learning, godard2019digging,feng2022disentangling,li2020enhancing,spencer2023kick, cheng2024adaptive}, different network architectures~\cite{zhou2021self,zhao2022monovit,lyu2020hr, xu2022attention} and leveraging additional constraints~\cite{guizilini2020semantically,li2023ldls,sun2023sc, schmied2023r3d3}. 
Our method falls in the self-supervised category. However, we estimate a true 3D representation from a single view, as opposed to the 2.5D representation produced by traditional depth estimation.

\par

\noindent \textbf{Semantic priors for depth estimation.} Previous depth estimation methods use semantic information to enhance 2D feature representations with different fusion strategies \cite{guizilini2020semantically,jung2021fine,li2023ldls,choi2020safenet}, or to remove dynamic objects \cite{casser2019depth,lee2021learning} during training. These methods utilize semantic information in the 2D representation space. On the contrary, we use semantic information to enhance 3D point representations and to guide our 3D spatial attention mechanism.

\noindent \textbf{Neural radiance fields.} 
Neural radiance fields (NeRFs)~\cite{mildenhall2020nerf,yu2021pixelnerf} learn a volumetric 3D representation of the scene from a set of posed input views. In particular, they use volumetric rendering in order to synthesize novel views by sampling volume density and color along a pixel ray. Recent multi-view reconstruction methods~\cite{wang2021neus,yariv2021volume,yu2022monosdf} take inspiration from this paradigm, reformulating the volume density function as a signed distance function for better surface reconstruction. These methods are focused on single-scene optimization using multi-view constraints, often representing the scene with the weights of a single MLP. 

To address the issue of generalization across scenes, Yu~\etal~\cite{yu2021pixelnerf} propose PixelNeRF to train a CNN image encoder across scenes that is used to condition an MLP, predicting volume density and color without multi-view optimization during inference. However, their approach is limited to small-scale and synthetic datasets.
Recently, Wimbauer~\etal~\cite{wimbauer2023behind} proposed  BTS, an extension of PixelNeRF to large-scale outdoor scenes. They omit the color prediction and, during training, use reference images to query color for a 3D point given a density field that represents the 3D scene geometry. While this simplification allows them to scale to large-scale outdoor scenes, their method falls short in predicting accurate geometry for occluded areas (\cref{fig:teaser}). We address this shortcoming by injecting fine-grained semantic knowledge and reasoning about semantic 3D context when querying the density field, leading to better shape recovery for occluded regions in particular.

\noindent \textbf{Scene as occupancy.} A recent line of work infers 3D scene geometry as voxelized 3D occupancy~\cite{cao2022monoscene,miao2023occdepth} from a single image. These works predict the occupancy and semantic class of each 3D voxel based on exhaustive 3D annotations. Therefore, these methods rely heavily on manually annotated datasets. Further, the predefined voxel resolution limits the fidelity of their 3D representation. In contrast, our method does not rely on labor-intensive manual annotations and represents the scene as a continuous density field.

\par
\noindent \textbf{Semantic priors for NeRFs.} 
Various works integrate semantic priors into NeRFs. While some utilize 2D semantic or panoptic semgentation~\cite{zhi2021place, fu2022panoptic,kundu2022panoptic}, others leverage 2D \ac{vl} features~\cite{kerr2023lerf,feng2022disentangling,peng2023openscene} and lift these into 3D space by distilling them into the NeRF. This enables the generation of 2D segmentation masks from new viewpoints, segmenting objects in 3D space, and discovering or removing particular objects from a 3D scene. While the aforementioned methods focus on the classical multi-view optimization setting, we instead focus on single-view input and leverage semantic priors to improve the 3D representation itself. 

\begin{figure*}[t]
\centering
% \vspace{-15pt}
\includegraphics[width=\linewidth]{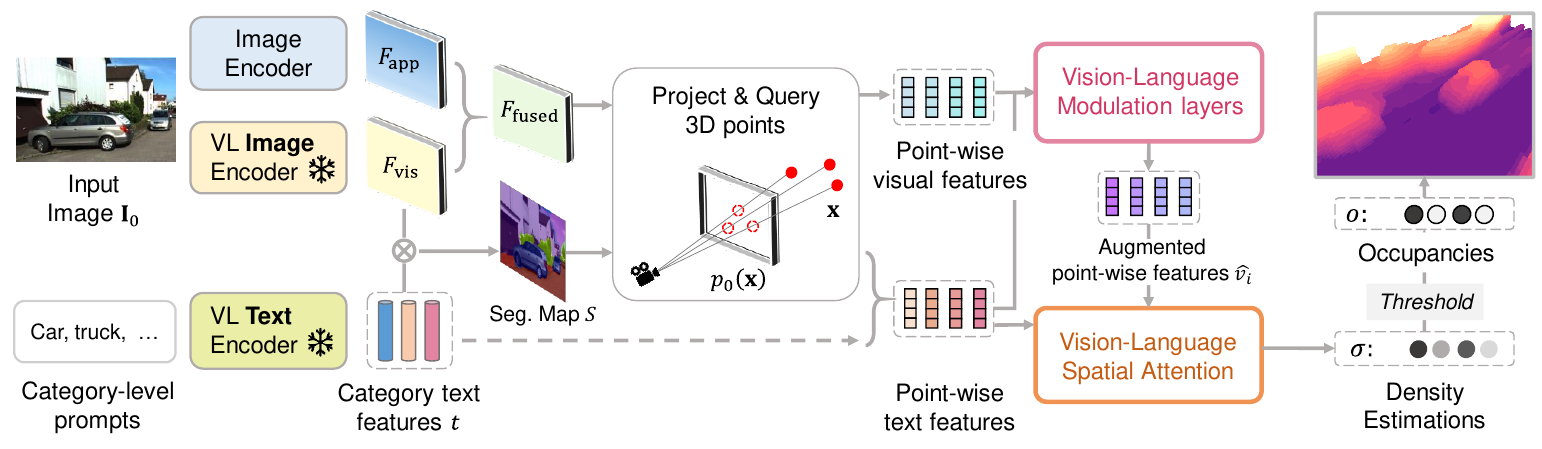}
% \vspace{-15pt}
\caption{\textbf{Overview.} Given an input image $\textbf{I}_{0}$, we use two image encoders to obtain features ($F_{\text{app}}$, $F_{\text{vis}}$), and fuse these into feature map $F_{\text{fused}}$. We further extract category-level text features and a segmentation map $S$. For a given 3D point set $\mathbf{X}$, we query the extracted features by projecting them onto the image plane yielding point-wise visual and text features. Next, the \ac{vl} modulation layers endow the point representation with fine-grained semantic information. Finally, the \ac{vl} spatial attention aggregates these point representations across the 3D scene, yielding density predictions aware of the 3D semantic context.
}
\label{fig:overview}
\end{figure*}

\section{Method}\label{sec:method}
\par
\noindent\textbf{Problem setup.} 
Given an input image $\mathbf{I}_{0}$, its corresponding intrinsics $K_{0} \in \mathbb{R}^{3\times 4}$ and pose $T_{0} \in \mathbb{R}^{4\times 4}$, we aim to reconstruct the full 3D scene by estimating the density for each 3D point $\mathbf{x} \in \mathbb{R}^{3}$ among point set $\mathbf{X}=\{\mathbf{x}_{i}\}^{M}_{i=1}$
{\begin{equation}
    \sigma_i = f(\mathbf{I}_{0}, K_{0}, T_{0}, \mathbf{X}, \theta),
\end{equation}}
where {the density $\sigma_i$ of point $\mathbf{x}_{i}$ is a function of the point set $\mathbf{X}$ and image $\mathbf{I}_{0}$ along with its camera intrinsic/extrinsics.} $f$ denotes the network and $\theta$ represents its parameters.
{The density $\sigma_i$ can be further transformed to the binary occupancy score $o_i \in \{0, 1\}$ with a predefined threshold $\tau$.}
During training (Sec.~\ref{sec:loss}), additional images $\mathbf{I}_{n}$ are incorporated with their corresponding intrinsics $K_{n}$ and extrinsics $T_{n}$, with $n\in \{1,\dots, N\}$, providing multiview supervision.

\par
\noindent\textbf{Overview.} 
We illustrate our method in Fig.~\ref{fig:overview}. Given an input image $\mathbf{I}_{0}$, we first extract image and \ac{vl} feature maps $F_\text{app}$ and $F_\text{vis}$. Next, we fuse the image and \ac{vl} features into a single feature map $F_\text{fused}$ and further utilize category-wise text features to compute a segmentation map $S$. We then use intrinsics $K_{0}$ to project the 3D point set $\mathbf{X}$ to the image plane and query $F_\text{fused}$, yielding point-wise visual features. In parallel, we retrieve point-wise text features by querying the segmentation map $S$ and looking up the corresponding category-wise text features $t$ for each 3D point. 
Given the point-wise visual and text features, we use our \ac{vl} modulation layers to augment the features with fine-grained semantic information. 
We aggregate the point-wise features with \ac{vl} spatial attention, adding 3D semantic context to the point-wise features. We guide the attention mechanism with the \ac{vl} text features. We finally predict a per-point density that is thresholded to yield the 3D occupancy map.

\subsection{Vision-Lanugage Modulation} \label{sec:vl_modu}
We detail how we extract point-wise visual features and how we augment the visual features with semantic information using our \ac{vl} modulation module.

\par
\noindent\textbf{Point-wise visual feature extraction.} 
Given the input image $\mathbf{I}_{0}$, we extract image features from standard image encoders~\cite{he2016deep} and \ac{vl} image encoder~\cite{li2022languagedriven}
\begin{equation} 
\begin{aligned}
    F_{\text{app}} &= f(\mathbf{I}_{0}, \theta_{\text{app}}), \\
    F_{\text{vis}} &= f(\mathbf{I}_{0}, \theta_{\text{vis}}),
\end{aligned}  
\end{equation}
where $F_{\text{app}}$ and $F_{\text{vis}}$ refer to the appearance features and \ac{vl} image features, respectively. We freeze the \ac{vl} image encoder weights $\theta_{\text{vis}}$ to retain the pre-trained semantics. We then fuse the features by concatenation followed by 2 convolutional layers, yielding $F_\text{fused}$. To obtain point-wise features for the 3D points $\mathbf{X}=\{\mathbf{x}_{i}\}^{M}_{i=1}$, we extract the fused feature $F_\text{fused}$ \textit{w.r.t.} the projected {2D} coordinates {$p_0{(\mathbf{x}_{i})}$} of each 3D point. Then, we combine each 3D point feature with a positional embedding $\gamma(\cdot)$ encoding its position in normalized device coordinates (NDC). We obtain the fused point-wise visual feature $ v_{i} \in \mathbb{R}^{1 \times C}$ for a 3D point $\mathbf{x}_{i}$
\begin{equation}
    v_{i} = \text{Concat}(F_\text{fused}(p_{0}(\mathbf{x}_{i})), \gamma(\mathbf{x}_i^{0})),
\end{equation}
where $\text{Concat}(\cdot,\cdot)$ is the feature concatenation operation, and $\mathbf{x}_i^{0}$ is the 3D position \textit{w.r.t.} $\mathbf{I}_0$'s coordinate. 

\par
\noindent\textbf{Point-wise text feature extraction.} As the text feature size does not align with the image space, it can not be extracted directly by querying projected 2D coordinates. To this end, we first derive the semantic category of each 3D point through 2D segmentations. Then we use it to associate text features with each 3D point.  
We utilize the category names from outdoor scenes \cite{cordts2016cityscapes} as prompts to the text encoder, yielding the text features $ t \in \mathbb{R}^{Q \times C}$, where $Q$ is the number of categories and $C$ is the feature channel. We then use the visual feature $F_\text{vis} \in \mathbb{R}^{H \times W \times C}$ from the \ac{vl} image encoder, to obtain the 2D semantic map by computing cosine similarity between \ac{vl} image and text features
\begin{equation}
    S = \argmax_{q \in \{1, \dots, Q\}} \frac{F_\text{vis} \otimes t^\top}{\|F_\text{vis}\| \|t\|}, \\
\end{equation}
where $\otimes$ denotes matrix multiplication, $S$ denotes the segmentation map by maximizing the similarity scores between \ac{vl} image and text features along the category dimension. We then compute the 
% category belongings 
semantic category for each 3D point by querying $S$ with projected 2D coordinates and then leverage it to obtain the text feature of each 3D point.
\begin{equation}
\begin{aligned}
    s_i &= S(p_0(\mathbf{x}_i)),\\
    g_{i}^{t} &= t(s_i),
\end{aligned}
\end{equation}
where $g_{i}^{t} \in \mathbb{R}^{C}$ is the text feature of the 3D point $\textbf{x}_{i}$.

\par
\noindent\textbf{\ac{vl} modulation layers.} 
To augment the 3D point features with rich semantics from both \ac{vl} image and text features, we propose the \ac{vl} modulation layers, which integrate both image feature $v_{i}$ and text representations $g_{i}^{t}$ of a 3D point $\textbf{x}_{i}$ for better scene geometry reasoning. 
The module is composed of $L=4$ modulation layers, each conducting modulation with multiplication operations
% the $l$-th layer conducts
\begin{equation}
    v_{i}^{l+1} = {\text{ReLU}}(\text{FC}(v_{i}^{l}) \odot \text{FC}(g_{i}^{t})),
\end{equation}
where $\text{FC}(\cdot)$ stands for the fully connected layer, $\odot$ is the element-wise product between features,  $\text{FC}(g_{i}^{t})$ denotes the text features encoded by a single fully-connected layer and is shared across different modulation layers. We set $v_{i}^{1} = v_{i}$ at the first modulation layer, and iterate through different layers. We utilize skip connections to inject the initial visual information $v_{i}^{1}$ after the modulated feature $v_{i}^{l'+1}$ at level $l'$, using concatenation followed by one fully-connected layer.
The output feature $\hat{v}_{i} = v_{i}^{L}$ denotes the 3D point feature of $\mathbf{x}_{i}$ augmented with rich image and text semantics.

\subsection{\ac{vl} Spatial Attention} \label{sec:vl_attn}
Next, we dive into our \ac{vl} spatial attention mechanism that aggregates the extracted point-wise features across the scene in a global-to-local fashion.
First, we combine the {whole set of point-wise visual features $\{\hat{v}_i\}_{i=1}^{M}$ and text features $\{g_{i}^{t}\}_{i=1}^{M}$ into $V \in \mathbb{R}^{M \times C'}$ and $C_{t} \in \mathbb{R}^{M \times C}$}. Then, we aggregate these features using a cross-attention operation in 3D space. Specifically, we leverage linear attention \cite{katharopoulos2020transformers, shen2021efficient} over appropriately split point sets to achieve memory-efficient spatial context reasoning.

\par
\noindent \textbf{Category-informed cross-attention.} We take the point-wise features $V$ as queries and values, and leverage text-based feature $C_{t}$ as the keys in the linear attention. Specifically, we project the features with fully connected layers to keys, queries, and values
\begin{equation}
\begin{aligned}
    F_{Q} &= f(V, \theta_{Q}), \\
    F_{K} &= f(C_t, \theta_{K}), \\
    F_{V} &= f(V, \theta_{V}), \\
\end{aligned}
\end{equation}
where all features are in $\mathbb{R}^{M \times \hat{C}}$, where $\hat{C}$ denotes the feature dimension before the attention. We then compute the global context score $G \in \mathbb{R}^{\hat{C} \times \hat{C}} $ by attending to the key and value features, and then correlating with query features by
\begin{equation}
\begin{aligned}
    G &= \text{Softmax}(F_K^\top) \otimes F_V, \\
    F_\text{final} &= \text{Softmax}(F_Q / \sqrt{D}) \otimes G,
\end{aligned}
\end{equation}
where $F_\text{final}$ denotes the spatially aggregated point-wise features. The density value is estimated by a single fully connected layer followed by a {Softplus$(\cdot)$} function
\begin{equation}
    \sigma = \text{Softplus}(\text{FC}(F_\text{final})).
\end{equation}

\par
\noindent \textbf{Reducing the memory footprint.} 
As the point features are sampled from the entire 3D space, simultaneously processing all point features can hit computational bottlenecks even with linear attention. 
To this end, we randomly split the initial points into chunks, to ensure that each chunk is identically distributed. Then we conduct the spatial point attention separately within each chunk and combine the density estimations afterward. As such, the attention can aggregate semantic point representations with both spatial awareness and efficiency.

\begin{figure*}
    \centering
    % \resizebox{0.9\linewidth}{!}{
    % Row for Input
    \subfloat{\begin{minipage}[c]{0.03\textwidth}
        \centering
        \footnotesize
        \rotatebox{90}{\makebox{Input}}
    \end{minipage}
    \hfill
    \begin{minipage}[c]{0.90\textwidth}
        \subfloat{\includegraphics[width=0.33\linewidth]{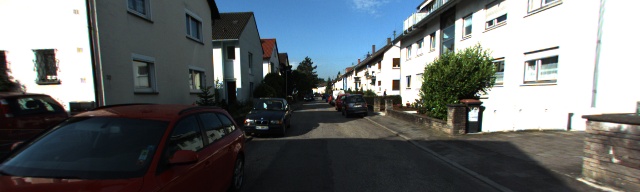}}
        \hfill
        \subfloat{\includegraphics[width=0.33\linewidth]{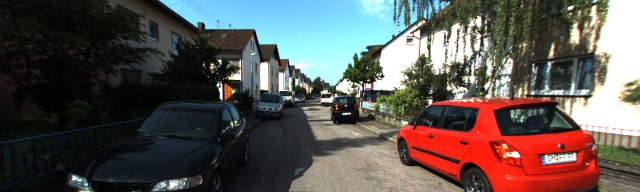}}
        \hfill
        \subfloat{\includegraphics[width=0.33\linewidth]{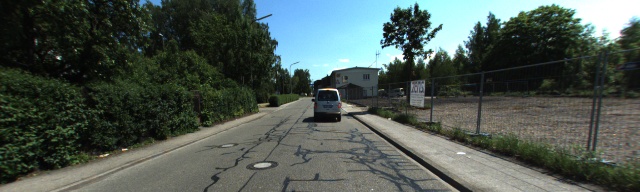}}
    \end{minipage}} \\ \vspace{5pt}

    % Monodepth2
    \subfloat{\begin{minipage}[c]{0.03\textwidth}
        \centering
        \footnotesize
        \rotatebox{90}{\makebox{Mono2 \cite{godard2019digging}}}
    \end{minipage}
    \hfill
    \begin{minipage}[c]{0.90\textwidth}
        \subfloat{\includegraphics[width=0.33\linewidth]{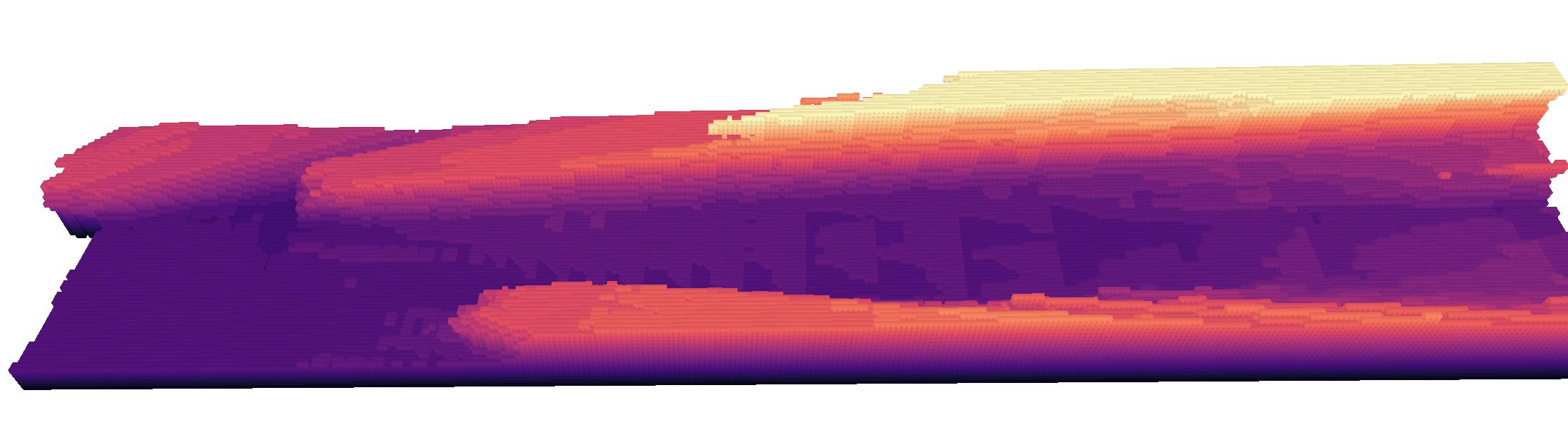}}
        \hfill
        \subfloat{\includegraphics[width=0.33\linewidth]{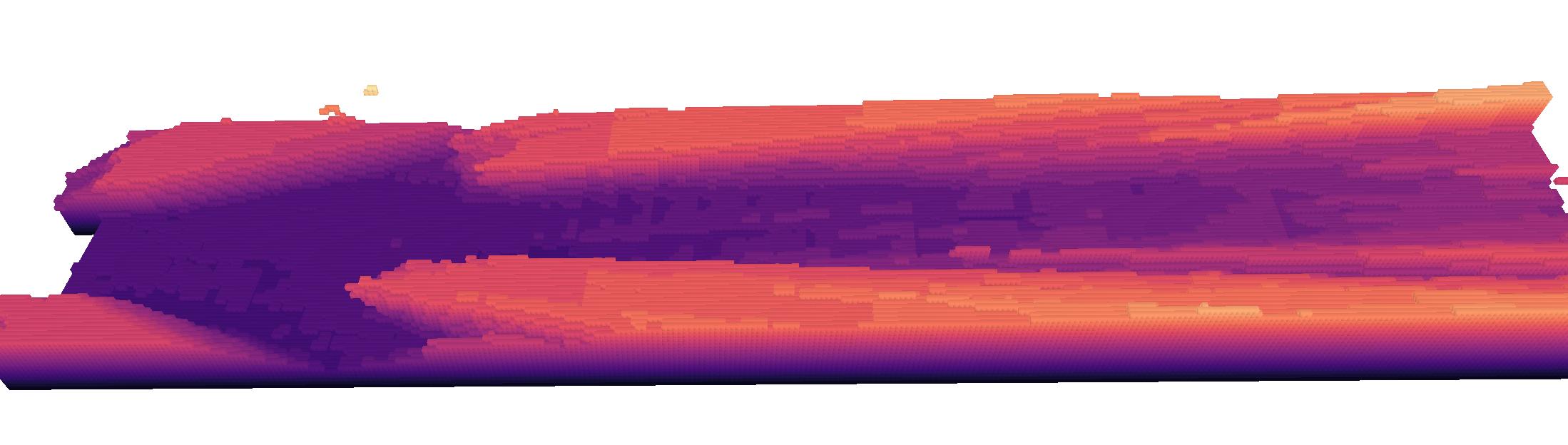}}
        \hfill
        \subfloat{\includegraphics[width=0.33\linewidth]{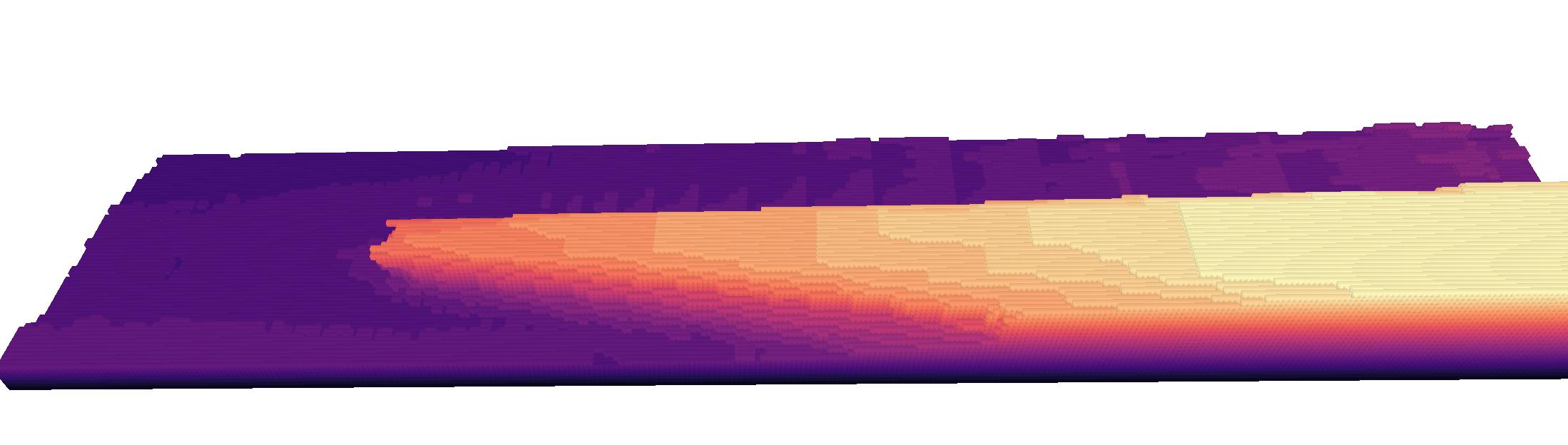}}
    \end{minipage} 
    } \\ \vspace{5pt}

    % Pixelnerf
    \subfloat{\begin{minipage}[c]{0.03\textwidth}
        \centering
        \footnotesize
        \rotatebox{90}{\makebox{PixelNeRF \cite{yu2021pixelnerf}}}
    \end{minipage}
    \hfill
    \begin{minipage}[c]{0.90\textwidth}
        \subfloat{\includegraphics[width=0.33\linewidth]{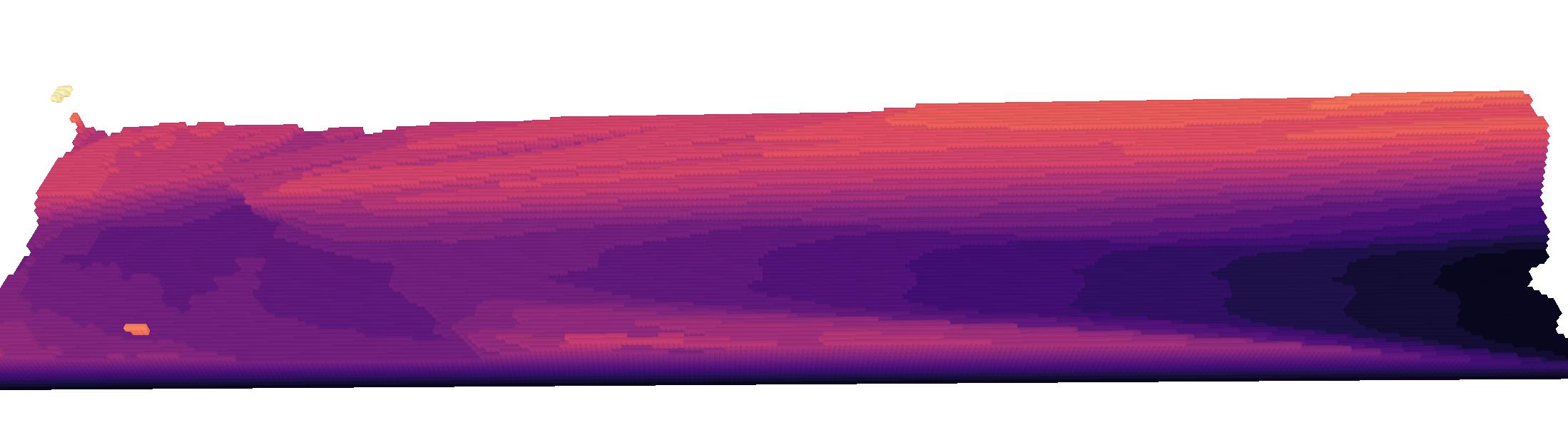}}
        \hfill
        \subfloat{\includegraphics[width=0.33\linewidth]{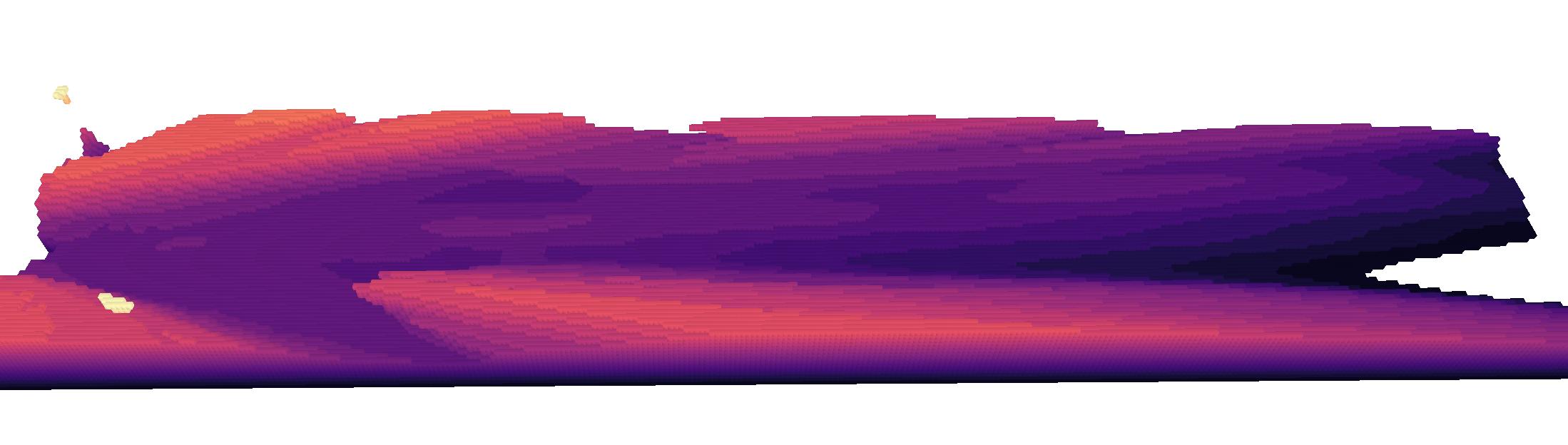}}
        \hfill
        \subfloat{\includegraphics[width=0.33\linewidth]{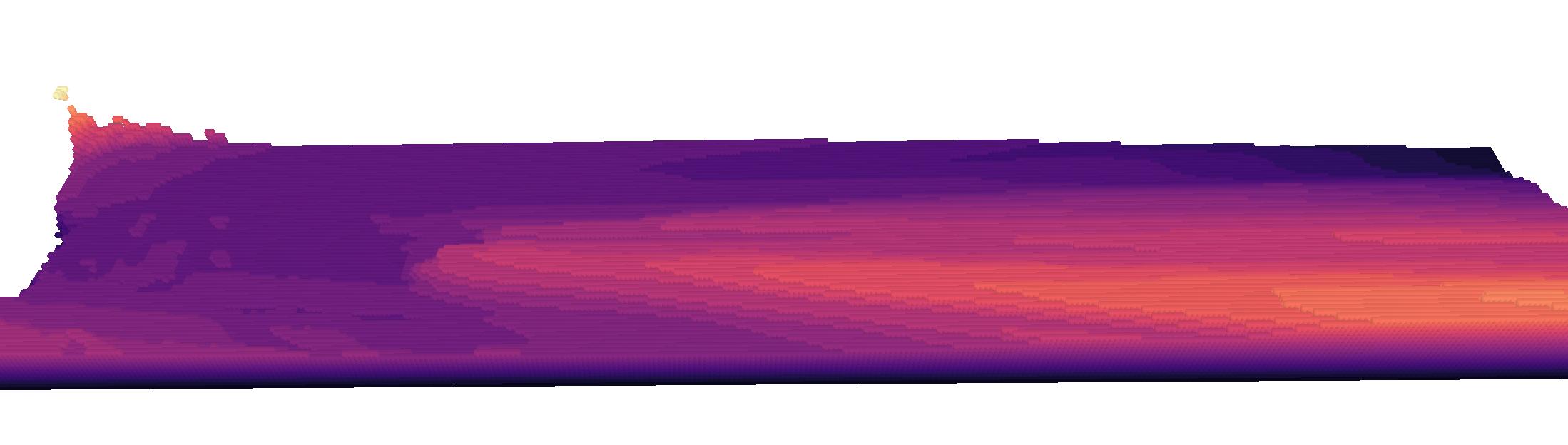}}
    \end{minipage} 
    } \\ \vspace{5pt}

    % BTS
    \subfloat{\begin{minipage}[c]{0.03\textwidth}
        \centering
        \footnotesize
        \rotatebox{90}{\makebox{BTS \cite{wimbauer2023behind}}}
    \end{minipage}
    \hfill
    \begin{minipage}[c]{0.90\textwidth}
        \subfloat{\includegraphics[width=0.33\linewidth]{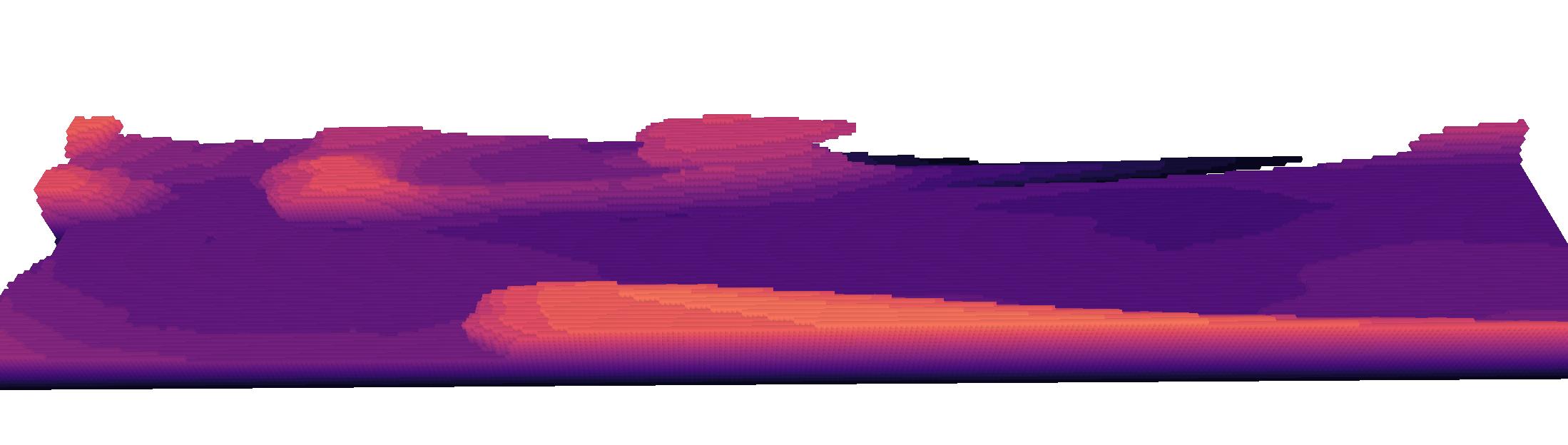}}
        \hfill
        \subfloat{\includegraphics[width=0.33\linewidth]{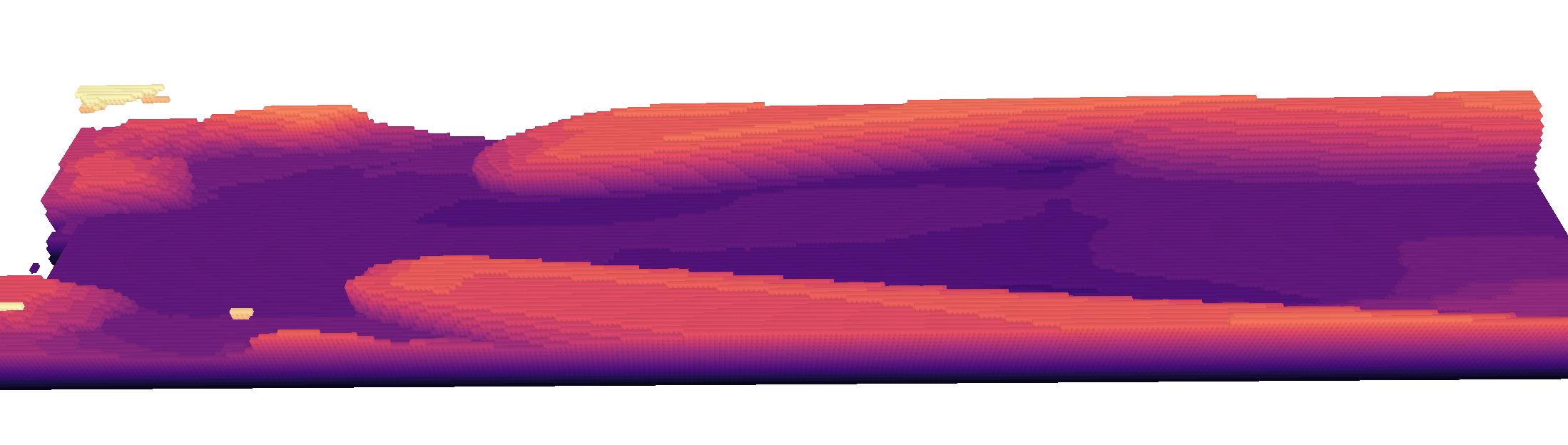}}
        \hfill
        \subfloat{\includegraphics[width=0.33\linewidth]{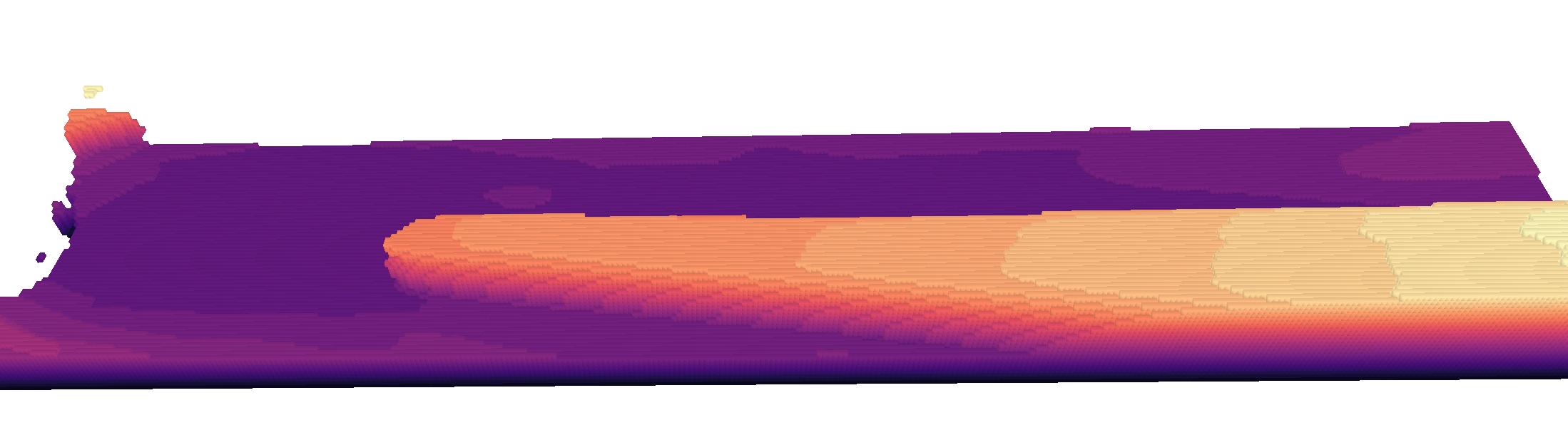}}
    \end{minipage} 
    } \\ \vspace{5pt}

    % BTS
    \subfloat{\begin{minipage}[c]{0.03\textwidth}
        \centering
        \footnotesize
        \rotatebox{90}{\makebox{\textbf{Ours}}}
    \end{minipage}
    \hfill
    \begin{minipage}[c]{0.90\textwidth}
        \subfloat{\includegraphics[width=0.33\linewidth]{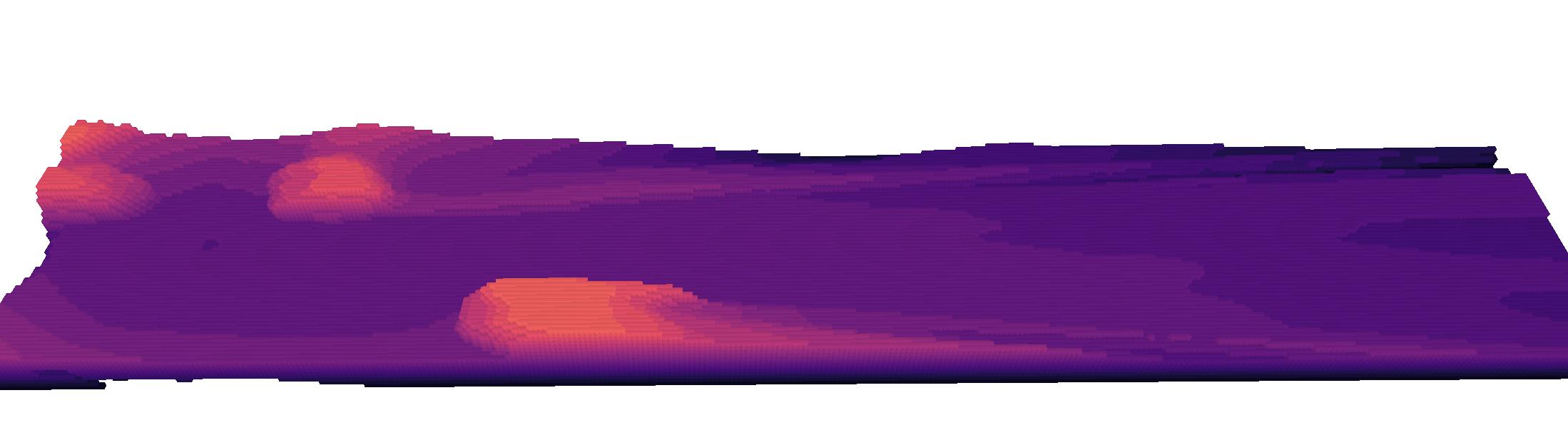}}
        \hfill
        \subfloat{\includegraphics[width=0.33\linewidth]{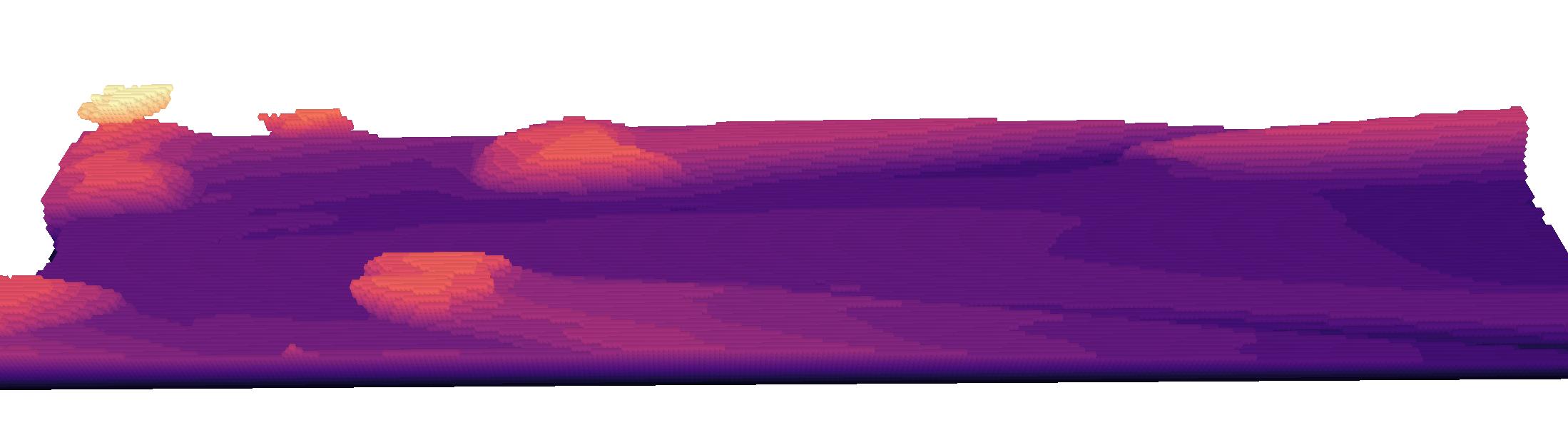}}
        \hfill
        \subfloat{\includegraphics[width=0.33\linewidth]{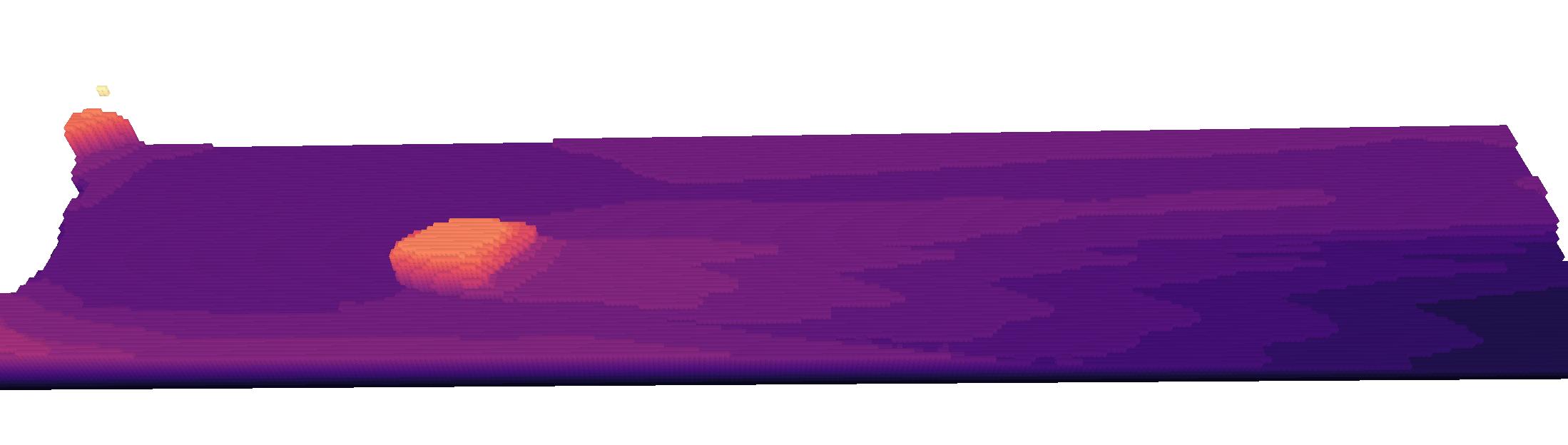}}
    \end{minipage} 
    }
% \vspace{-5pt}
\caption{\textbf{Qualitative comparisons on KITTI-360 dataset.} We illustrate the scene reconstructions as voxel grids, where the camera is on the left side and points to the right along the $z$-axis. A lighter voxel color indicates higher voxel positions. Compared to previous methods that struggle with corrupted and trailing shapes, our method produces faithful scene geometry, especially for occluded areas.}
\label{fig:scene_recon}
\end{figure*}

\subsection{Training Process} \label{sec:loss}
Our method achieves self-supervision by computing the photometric loss between the reconstructed and target colors. We extract the 2D feature map of $\textbf{I}_{0}$ to get point-wise representations, then partition all images $\{\textbf{I}_{0}\} \cup \{\textbf{I}_{n}\}_{n=1}^{N}$ into a source set $N_{\text{source}}$ and a loss set $N_{\text{loss}}$ following previous practice \cite{wimbauer2023behind}. We render the RGB of $N_{\text{loss}}$ from the corresponding colors of $N_{\text{source}}$ similar to the self-supervised depth estimation \cite{godard2019digging}. Instead of resorting to the whole image, we perform patch-based image supervision \cite{wimbauer2023behind} to reduce memory footprints. For each pixel on a patch, we sample the 3D points $\mathbf{x}_{i}$ on its back-projected ray and conduct density estimation.
Let $\mathbf{x}_{i}$ and $\mathbf{x}_{i+1}$ be the adjacent sampled pixels on a ray, 
we calculate the RGB information by volume rendering the \textit{sampled} color \cite{wimbauer2023behind}
\begin{equation}
    \alpha_i=\exp \left(1-\sigma_{\mathbf{x}_i} \delta_i\right) \quad T_i=\prod_{j=1}^{i-1}\left(1-\alpha_j\right),
\end{equation}
\begin{equation}
\hat{d}=\sum_{i=1}^S T_i \alpha_i d_i \quad
\hat{c}_k=\sum_{i=1}^S T_i \alpha_i c_{\mathbf{x}_i, k},
\end{equation}
where $\delta_i$ denotes the distance between adjacent sampled points $\mathbf{x}_{i}$ and $\mathbf{x}_{i+1}$ along the ray, and $\alpha_i$ refer to the probability that the ray ends between $\mathbf{x}_{i}$ and $\mathbf{x}_{i+1}$. Note that the color $c_{\mathbf{x}_i, k} = \mathbf{I}_{k}(p_k(\mathbf{x}_{i}))$ is the sampled RGB value from the view $k$ in the source set $N_{\text{source}}$, to obtain a better geometry. $\hat{d}$ and $\hat{c}_k$ represent the terminating depth and the rendered color.
\par
As we sample the pixels in a patch-wise manner during training, the rendered RGB and depth are also organized in patches.  Let $\hat{P}_{k}$ as the rendered patch from view $k$ in the source $N_{\text{source}}$, $P$ as the supervisory patch from $N_{\text{loss}}$, and $d'$ as the patch depth of $P$, the loss function is defined {following previous methods \cite{wimbauer2023behind, godard2019digging}}
\begin{equation}
    \mathcal{L} = \mathcal{L}_{ph} + \lambda_e \mathcal{L}_e,
\end{equation}
where $\lambda_e = 10^{-3}$, $\mathcal{L}_{ph}$ and $\mathcal{L}_{e}$ are photometric loss and edge-aware smoothness loss \cite{godard2019digging} on patches 
\begin{equation}
    \mathcal{L}_{ph}=\min _{k \in N_{\text {render }}}\left(\lambda_{1} \text{L1}\left(P, \hat{P}_{k}\right)+\lambda_{2} \text{SSIM}\left(P, \hat{P}_{k}\right)\right),
\end{equation}
\begin{equation}
    \mathcal{L}_{e}=\left|\delta_x d_{i}' \right| e^{-\left|\delta_x P\right|}+\left|\delta_y d_{i}' \right| e^{-\left|\delta_y P\right|},
\end{equation}
where $\lambda_1=0.15$ and $\lambda_1=0.85$, $\delta_x, \delta_y$ denotes the gradient along the horizontal and vertical directions. 
\section{Experiments} \label{sec:exp}
To demonstrate the effectiveness of our proposed method, we compare with existing works~\cite{zhao2022monovit,godard2019digging,yu2021pixelnerf,wimbauer2023behind} in single-view scene reconstruction, including both depth estimation~\cite{godard2019digging} and radiance field based methods~\cite{yu2021pixelnerf,wimbauer2023behind}. We evaluate both the 3D scene (Sec. \ref{sec:exp_kt360_scene}) and object (Sec. \ref{sec:exp_kt360_obj}) reconstruction results on the KITTI-360 dataset in short and long ranges. 
We conduct extensive ablations (Sec. \ref{sec:exp_ablation}) to verify the effectiveness of each contribution, compare our method with existing semantic feature fusion techniques, as well as evaluate our method's performance with the broader supervision range. Moreover, we demonstrate our method's zero-shot generalization ability in Sec. \ref{sec:exp_generalization}.

\subsection{Datasets}
We use the KITTI-360 \cite{liao2022kitti} dataset for training and evaluation, as it captures static scenes using cameras deployed with wide baselines, \ie, two stereo cameras, and two side cameras (fisheye cameras) at each timestep, which facilitate learning full 3D shapes by self-supervision. During the training phase, we use all cameras from two time steps, \ie 8 cameras in total, to train our model. We use an input resolution of 192$\times$640 and choose the left stereo camera from the first time step to extract the image features. We split all cameras randomly into $N_{\text{render}}$ and $N_{\text{loss}}$ for sampling colors and loss computation, as is illustrated in Sec. \ref{sec:loss}.
Furthermore, we use the DDAD \cite{guizilini20203d} dataset to evaluate the zero-shot generalization capability of the models trained on KITTI-360. We select testing sequences with more than 50 images and use 384$\times$640 image resolution.

% scene recon with color
\begin{table}
\centering
\footnotesize
\begin{tabular}{llccc}
\toprule[1pt]
\multicolumn{2}{c}{Method}  & O${_\text{acc}}\uparrow$ & IE$_{\text{acc}}\uparrow$ & IE$_{\text{rec}}\uparrow$ \\
\midrule
\multirow{5}{*}{4-20m} & Monodepth2 \cite{godard2019digging} & \nd 0.90 & n/a & n/a \\
& Monodepth2 \cite{godard2019digging} + $4m$ & \nd 0.90 & 0.59 & \nd{0.66} \\
& PixelNeRF \cite{yu2021pixelnerf} & \rd 0.89 & \rd0.62 & 0.60 \\
& BTS \cite{wimbauer2023behind} & \fs{0.92} & \nd{0.66} & \rd0.64 \\
& \textbf{Ours} & \fs{0.92} & \fs{0.70} & \fs{0.72} \\
\midrule
\multirow{5}{*}{4-50m} & Monodepth2 \cite{godard2019digging} & \rd0.82 & n/a & n/a \\
&Monodepth2 \cite{godard2019digging} + $4m$ & 0.81 & 0.54 & \fs{0.76} \\
& PixelNeRF \cite{yu2021pixelnerf} & \rd0.82 & \rd0.56 & \rd0.68 \\
& BTS \cite{wimbauer2023behind} & \nd{0.84} & \nd{0.61} & {0.53} \\
& \textbf{Ours} & \fs{0.86} & \fs{0.63} & \nd{0.73} \\
\bottomrule[1pt]
\end{tabular}
% \vspace{-5pt}
\caption{\textbf{Comparison of \textit{scene} reconstruction on KITTI-360}. Our method achieves the best overall performance in both the near and far evaluation range.}\label{tab:scene_recon}
\end{table}
% \vspace{-10pt}

\subsection{Evaluation}
We follow the experimental protocol in~\cite{wimbauer2023behind} to evaluate 3D occupancy prediction. Specifically, we sample 3D grid points on 2D slices parallel to the ground plane. For KITTI-360, we report scores within distance ranges $[4, 20]$ and $[4, 50]$ meters, where the latter provides a more challenging evaluation scenario. For ground-truth generation, we follow~\cite{wimbauer2023behind} and accumulate 3D LiDAR points across time, and set 3D points lying out of all depth surfaces as unoccupied, otherwise set to occupied. Unlike~\cite{wimbauer2023behind}, which accumulate only 20 LiDAR sweeps often leading to inaccurate occluded scene geometry, we accumulate up to 300 LiDAR frames. We also provide results with a 20-frame accumulated ground truth in the supplementary material for reference.
For the DDAD dataset, we accumulate up to 100 LiDAR frames due to the limited sequence length and evaluate in the $[4, 50]$ meters range.
\par
\noindent \textbf{Metrics.} We adopt the evaluation metrics in~\cite{wimbauer2023behind} and measure overall reconstruction (O$_{\text{acc}}$) and occluded reconstruction (IE$_{\text{acc}}$, IE$_{\text{rec}}$) accuracies. Specifically, O$_{\text{acc}}$ computes the accuracy between the prediction and the ground truth in the full area of the evaluation range, thus reflecting the overall performance of the reconstruction. IE$_{\text{acc}}$ computes the accuracy of the invisible areas specifically, \ie without direct visual observation in $\mathbf{I}_{0}$. IE$_{\text{rec}}$ computes the recall of both invisible and empty areas, which evaluates the reconstruction of the occluded empty space. The three metrics focus on different aspects of the reconstruction quality. 
\par
\noindent \textbf{Scene- and object-level evaluation.} In addition to evaluating the performance of the whole scene, we focus on object reconstruction in particular because they are of particular interest compared to \eg the road plane.
To this end, we manually annotate the object areas in the ground-truth occupancy maps and compute the evaluation metrics on these object areas. We refer the reader to the supplementary material for details. 

\subsection{Implementation Details}
We implement our method using Pytorch \cite{paszke2017automatic} and train it on NVIDIA Quadro RTX 6000 GPUs. The appearance network is similar to \cite{godard2019digging} with pre-trained weights on ImageNet \cite{deng2009imagenet}. We adopt LSeg \cite{li2022languagedriven} as the visual-language network and freeze its parameters. The model is trained using Adam \cite{kingma2014adam} optimizer with a learning rate of $10^{-4}$ for 25 epochs, which is reduced to $10^{-5}$ after 120k iterations. During the training phase, we sample 4096 patches across loss set $N_\text{loss}$, each patch contains $8\times 8$ pixels. We further sample 64 points along each ray following \cite{wimbauer2023behind}.

\subsection{Scene Reconstruction}\label{sec:exp_kt360_scene}
We compare our method with recent single-view scene reconstruction methods using self-supervision \cite{godard2019digging,yu2021pixelnerf,wimbauer2023behind}. Specifically, we train Monodepth2 \cite{godard2019digging} on our benchmark as the base depth estimation method. Since the depth network cannot infer occluded geometry, we use a handcraft criterion to set areas $4m$ behind the depth surface as empty space (Monodepth2 + $4m$). We also compare our method with the NeRF-based methods \cite{yu2021pixelnerf, wimbauer2023behind} following the same training protocol. As shown in the Tab. \ref{tab:scene_recon}, compared to previous methods, our method achieves both the best overall performance (O$_{\text{acc}}$) and the best occluded area reconstruction (IE$_{\text{acc}}$, IE$_{\text{rec}}$).
Note that Monodepth2 (+ $4m$) also yields a competitive performance in terms of invisible and empty space reconstruction (IE$_{\text{rec}}$), but it relies on hand-crafted criteria that cannot learn true 3D in the scene. Qualitative comparisons are shown in Fig. \ref{fig:scene_recon}. We show the reconstructed occupancy grids, where the camera is on the left side and points to the right along the $z$-axis within [4, 50m] range.
Our method demonstrates obvious qualitative superiority in reasoning occluded object shapes against the inherent ambiguity. Notably, it substantially reduces the trailing effects.

\begin{table}
\centering
\footnotesize
\begin{tabular}{llccc}
\toprule[1.0pt]
\multicolumn{2}{c}{Method}  & O${_\text{acc}}\uparrow$ & IE$_{\text{acc}}\uparrow$ & IE$_{\text{rec}}\uparrow$ \\
\midrule
\multirow{5}{*}{4-20m} & Monodepth2 \cite{godard2019digging} & 0.69 & n/a & n/a \\
& Monodepth2 \cite{godard2019digging} + $4m$ & \rd0.70 & \nd 0.53 & \rd{0.52} \\
& PixelNeRF \cite{yu2021pixelnerf} & 0.67 & \nd0.53 & 0.49 \\
& BTS \cite{wimbauer2023behind} & \nd{0.79} & \fs{0.69} & \nd{0.60} \\
& \textbf{Ours} & \fs{0.80} & \fs{0.69} & \fs{0.70} \\

\midrule
\multirow{5}{*}{4-50m} & Monodepth2 \cite{godard2019digging} & 0.65 & n/a & n/a \\
&Monodepth2 \cite{godard2019digging} + $4m$ & \rd0.68 & 0.48 & \nd{0.59} \\
& PixelNeRF \cite{yu2021pixelnerf} & 0.66 & \rd0.56 & \rd0.58 \\
& BTS \cite{wimbauer2023behind} & \nd{0.72} & \nd{0.61} & {0.48} \\
& \textbf{Ours} & \fs{0.75} & \fs{0.64} & \fs{0.68} \\
\bottomrule[1.0pt]
\end{tabular}
\vspace{-5pt}
\caption{\textbf{Comparison of \textit{object} reconstruction on KITTI-360}. Our method outperforms other methods in all metrics in both the short and long evaluation range.}\label{tab:obj_recon}
\end{table}
% \vspace{-10pt}

\subsection{Object Reconstruction}\label{sec:exp_kt360_obj}
\begin{figure}[t]
\centering
% \vspace{-12pt}
\subfloat{
    \begin{minipage}[c]{0.02\linewidth}
        \centering
        \scriptsize
        \rotatebox{90}{\makebox{Input}}
    \end{minipage}
    \begin{minipage}[c]{0.90\linewidth}
        \centering
        \includegraphics[width=1\linewidth]{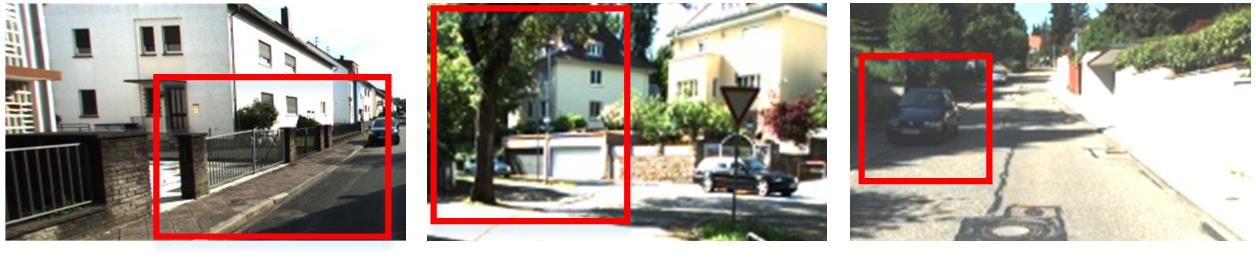}
    \end{minipage}
} \vspace{-3pt}\\
\subfloat{
    \begin{minipage}[c]{0.02\linewidth}
        \centering
        \scriptsize
        \rotatebox{90}{\makebox{Mono2\cite{godard2019digging}}}
    \end{minipage}
    \begin{minipage}[c]{0.90\linewidth}
        \centering
        \includegraphics[width=1\linewidth]{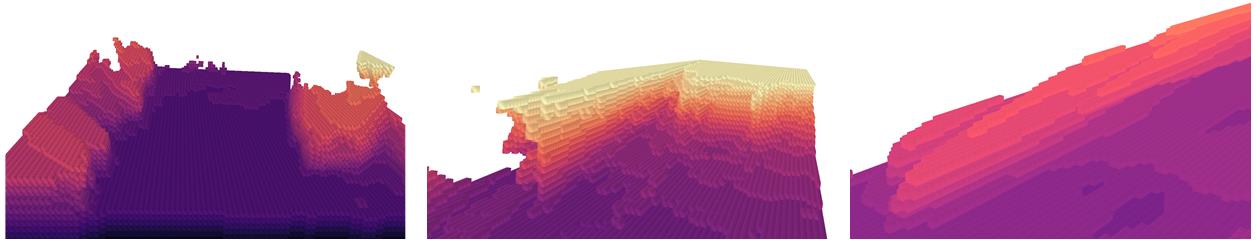}
    \end{minipage}
} \vspace{0.5pt}\\
\subfloat{
    \begin{minipage}[c]{0.02\linewidth}
        \centering
        \scriptsize
        \rotatebox{90}{\makebox{PixelNeRF \cite{yu2021pixelnerf}}}
    \end{minipage}
    \begin{minipage}[c]{0.90\linewidth}
        \centering
        \includegraphics[width=1\linewidth]{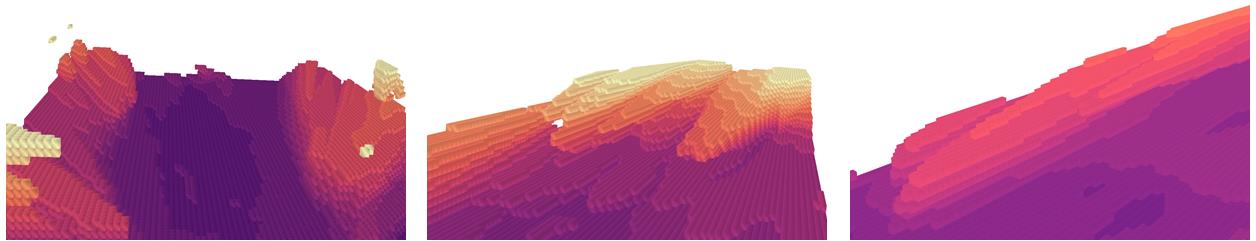}
    \end{minipage}
} \vspace{0.5pt}\\
\subfloat{
    \begin{minipage}[c]{0.02\linewidth}
        \centering
        \scriptsize
        \rotatebox{90}{\makebox{BTS \cite{wimbauer2023behind}}}
    \end{minipage}
    \begin{minipage}[c]{0.90\linewidth}
        \centering
        \includegraphics[width=1\linewidth]{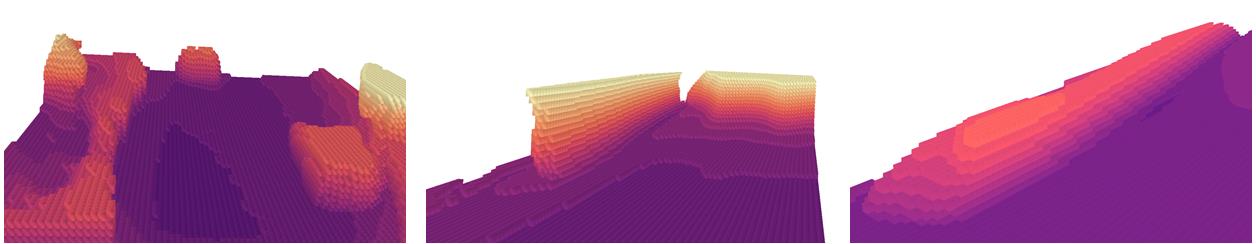}
    \end{minipage}
} \vspace{0.5pt}\\
\subfloat{
    \begin{minipage}[c]{0.02\linewidth}
        \centering
        \scriptsize
        \rotatebox{90}{\makebox{\textbf{Ours}}}
    \end{minipage}
    \begin{minipage}[c]{0.90\linewidth}
        \centering
        \includegraphics[width=1\linewidth]{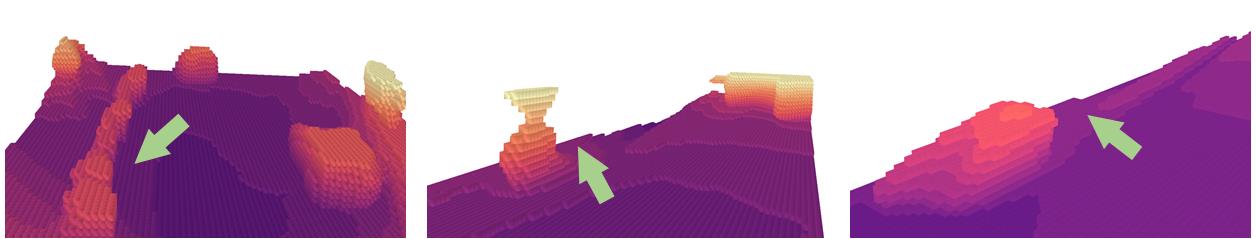}
    \end{minipage}
}\vspace{0.5pt} \\
% \vspace{-5pt}
\caption{\textbf{Object reconstruction in the KITTI-360 dataset \cite{liao2022kitti}.} From left to right: Reconstructions of the fence, tree, and car. Our method produces more faithful object geometries, in particular in occluded areas and for various semantic categories. 
}
% \vspace{-10pt}
\label{fig:obj_recon}
\end{figure}

We evaluate the object reconstruction performance by computing the metrics within the annotated object areas.
As shown in Tab. \ref{tab:obj_recon}, our method achieves competitive or better results in the [4, 20m] evaluation range. Meanwhile, it achieves obvious improvements for all metrics in the [4, 50m] range, demonstrating its effectiveness in reasoning ambiguous geometries away from the camera origin. We further show reconstructions of different categories in Fig. \ref{fig:obj_recon}. Our method generates faithful object shapes with reasonable estimates of occluded geometry for different categories including fences, trees, cars, \etc, showing clear improvements over existing methods.

\subsection{Ablation Studies}\label{sec:exp_ablation}
We evaluate the effectiveness of each contribution by separately ablating the \ac{vl} modulation (Sec. \ref{sec:abla_vl_modu}) and the \ac{vl} spatial attention (Sec. \ref{sec:abla_vl_attn}). 
We further compare with existing techniques injecting semantics from related tasks (Sec. \ref{exp:abla_cate_guidance}). Moreover, we show our method's improvement under a broader supervision range (Sec. \ref{sec:abla_better_sup}).

\subsubsection{Ablation study on VL Modulation}\label{sec:abla_vl_modu}
We evaluate the effectiveness of \ac{vl} modulation by adding different components upon baseline method \cite{wimbauer2023behind}, which uses only appearance feature $F_{\text{app}}$ from the generic backbone \cite{he2016deep}. In Tab.~\ref{tab:abla_vl_modulation}, we enhance the baseline by incorporating the fused \ac{vl} image features ($F_{\text{fused}}$) and injecting image/text semantics with the \ac{vl} Modulation (VL-Mod.). We find that merely using the fused feature $F_{\text{fused}}$ from the \ac{vl} image encoder does not contribute to overall improvement. As a comparison, the proposed VL Modulation significantly improves the performances by properly interacting with image and text features.

% vl modulation with "checkbox"
\begin{table}[t]
\centering
\footnotesize
\setlength{\tabcolsep}{4.2pt}
% \scalebox{0.78}{
\begin{tabular}{c c c c c c c c c}
\toprule[1.0pt] 
\multirow{2}{*}{$F_{\text{app}}$} & \multirow{2}{*}{$F_{\text{fused}}$} & \multirow{2}{*}{VL-Mod.} & \multicolumn{3}{c}{Scene Recon.} & \multicolumn{3}{c}{Object Recon.} \\
\cline{4-9}
& & & O$_{\text{acc}}$ & IE$_{\text{acc}}$ & IE$_{\text{rec}}$ & O$_{\text{acc}}$ & IE$_{\text{acc}}$ & IE$_{\text{rec}}$ \\
\midrule
\checkmark & & & \nd0.84 & \nd 0.60 & \rd 0.53 &  \nd 0.72& \nd 0.61 & \nd 0.48 \\
 & \checkmark &  &\nd 0.84 & \nd 0.60 & \nd 0.55 &  \nd 0.72& \nd 0.61 & \nd 0.48 \\
&\checkmark & \checkmark & \fs{0.85} & \fs{0.63} & \fs{0.64} & \fs{0.73} & \fs{0.63} & \fs{0.59} \\

\bottomrule[1.0pt]
\end{tabular}
% }
\vspace{-5pt}
\caption{\textbf{Ablations study on \ac{vl} modulation.} We report the performance in the [4, 50m] range. Naively introducing the \ac{vl} image feature does not improve performance. Our VL modulation correlating the image and text features yields the best scores.}\label{tab:abla_vl_modulation}
\end{table}

\begin{table}[t]
\centering
\footnotesize
\setlength{\tabcolsep}{1.5pt}
% \scalebox{0.80}{
\begin{tabular}{@{}c c c c c c c c c c c@{}}
\toprule[1.0pt] 
% \hline
\multirow{2}{*}{$F_{\text{app}}$} & \multirow{2}{*}{$F_{\text{fused}}$} & \multirow{2}{*}{VL-Mod.} & \multirow{2}{*}{Attn.} & \multirow{2}{*}{VL-Attn.} & \multicolumn{3}{c}{Scene Recon.} & \multicolumn{3}{c}{Object Recon.} \\
\cline{6-8} \cline{9-11}
& & & & & O$_{\text{acc}}$ & IE$_{\text{acc}}$ & IE$_{\text{rec}}$ & O$_{\text{acc}}$ & IE$_{\text{acc}}$ & IE$_{\text{rec}}$ \\
\midrule
% \hline
\checkmark & & & & & \rd 0.84 & \rd 0.60 & 0.53 &  0.72& \rd 0.61 & 0.48 \\
\checkmark &  & & \checkmark & & \nd 0.85 & \nd 0.61 & 0.60 &  \rd 0.73& \rd 0.61 & 0.56 \\ \midrule
 & \checkmark &  &\checkmark & & \nd 0.85 & \nd 0.61 & \rd  0.67 &  0.72& 0.60 & \rd 0.62 \\
 & \checkmark & & & \checkmark & \nd 0.85 & \rd 0.60 & 0.66 &  \rd 0.73& \nd 0.62 & 0.61 \\
 & \checkmark & \checkmark & \checkmark & & \fs{0.86} & \fs{0.63} & \fs{0.75} & \nd {0.74} & \nd {0.62} & \fs{0.73} \\
 & \checkmark & \checkmark & & \checkmark & \fs{0.86} & \fs{0.63} & \nd {0.73} & \fs{0.75} & \fs{0.63} & \nd  {0.68} \\

\bottomrule[1.0pt]
% \hline

\end{tabular}
% }
\vspace{-5pt}
\caption{\textbf{Ablation study on spatial attention.} We report the performance in the [4, 50m] range. The spatial attention improves in each variant by enabling the 3D context awareness. Combining the \ac{vl} features with spatial attention yields the best performance.}\label{tab:abla_vl_attention}
\end{table}
% \vspace{-15pt}

\subsubsection{Ablation Study on Spatial Attention}\label{sec:abla_vl_attn}
{In Tab. \ref{tab:abla_vl_attention}, we provide experimental support for the effectiveness of using spatial context, both via spatial attention and \ac{vl} spatial attention. We first add spatial attention over image appearance features only (row: 1$\rightarrow$2). Without the \ac{vl} features, aggregating spatial context still yields notable improvement. We further show that the introduction of semantics via \ac{vl} modulation (VL-Mod) helps independent of the spatial aggregation mechanism (row 3$\rightarrow$5, row: 4$\rightarrow$6), underpinning that adding VL text features outperforms using VL image features only. When combining both \ac{vl} modulation and spatial attention mechanisms, we achieve the best performances (rows 5, 6). Additionally, we observe a small albeit notable gain when also injecting VL features into the spatial aggregation (VL-Attn), however, mainly improving details are not captured in current metrics. Please refer to the supplementary material for details.
}

\begin{table}[t]
\centering
\setlength{\tabcolsep}{5.5pt}
\footnotesize
% \scalebox{0.8}{
\begin{tabular}{l l c c c}
\toprule[1.0pt]
{Semantics} & {Method} & O$_{\text{acc}}$ & IE$_{\text{acc}}$ & IE$_{\text{rec}}$ \\
\midrule
-- & Baseline & \nd 0.84 & \rd 0.60 &\rd 0.52\\ % \hline

\hline		
\multirow{3}{*}{Semantic Feat.}  & Plain Fusion &\nd 0.84 & \nd 0.61& 0.51\\
& 2D Fusion - LDLS \cite{li2023ldls} &\nd0.84  &\rd 0.60 & \nd 0.55\\
& 2D Fusion - SAFENet \cite{choi2020safenet} &\nd0.84 &\rd 0.60 &0.51\\
		
\hline
\multirow{2}{*}{VL Feat.} 
& Fusing \ac{vl} image feature & \nd{0.84} & \rd{0.60}& \nd{0.55}\\
& \textbf{Ours} & \fs{0.86} & \fs{0.63}& \fs{0.73}\\
\bottomrule[1.0pt]
\end{tabular}
% }
% \vspace{-5pt}
\caption{\textbf{Comparison with semantic feature fusion techniques.} We compare with 2D feature fusion techniques from semantic-guided depth estimation \cite{li2023ldls,choi2020safenet}. Our method achieves the best performance with visual and language semantic enhancement and 3D context awareness.
}
\end{table}

\subsubsection{Comparison with Other Semantic Guidances}\label{exp:abla_cate_guidance}
As semantic cues are vital in other tasks such as depth estimation, we investigate whether the techniques used in the related literature~\cite{li2023ldls, choi2020safenet} are useful for single-view scene reconstruction.
We use the pre-trained DPT \cite{ranftl2021vision} semantic network to provide pre-trained semantic features, and incorporate different feature fusion techniques \cite{li2023ldls, choi2020safenet} to our density prediction pipeline. As shown in Tab.~\ref{tab:abla_dist}, we find that conducting 2D feature fusion does not lead to notable improvements over the baseline, which is consistent even with incorporating \ac{vl} image feature fusion.
% \ac{vl} $F_{\text{fused}}$ fusion in Sec. \ref{sec:abla_vl_modu}. 
However, by appropriately interacting with image and text features with 3D context awareness, our method outperforms related techniques by a notable margin.

% \vspace{10pt}
\begin{table}[t]
\centering
\footnotesize
\setlength{\tabcolsep}{3.5pt}
% \scalebox{0.75}{
\begin{tabular}{l l c c c c c c}
\toprule[1.0pt]
\multirow{2}{*}{Supervision}& \multirow{2}{*}{Method} & \multicolumn{3}{c}{Scene Recon.} & \multicolumn{3}{c}{Object Recon.} \\
\cline{3-8}
 & & O$_{\text{acc}}$ & IE$_{\text{acc}}$ & IE$_{\text{rec}}$ & O$_{\text{acc}}$ & IE$_{\text{acc}}$ & IE$_{\text{rec}}$ \\
\midrule
\multirow{2}{*}{1$s$ in the future} & BTS \cite{wimbauer2023behind}& \nd 0.84 & \nd 0.61 & \nd 0.53 & \nd  0.72& \nd 0.61 & \nd 0.48 \\
& \textbf{Ours} & \fs{0.86} & \fs{0.63} & \fs{0.73} & \fs{0.75} & \fs{0.64} & \fs{0.68} \\
\hline
\multirow{2}{*}{1-4$s$ in the future} & BTS \cite{wimbauer2023behind}& \nd 0.86 & \nd 0.68 & \nd 0.72 & \nd 0.74 &\nd  0.64 &\fs{0.75} \\
& \textbf{Ours}  & \fs{0.87} & \fs{0.69} & \fs{0.77} & \fs{0.78} & \fs{0.68} & \fs{0.75} \\
\bottomrule[1.0pt]
\end{tabular}
% }
\vspace{-5pt}
\caption{\textbf{Evaluation of different supervisory ranges.} We report the performance in [4, 50m] range. Our method with the standard supervision range (1$s$ in the future) yields comparable performance with \cite{wimbauer2023behind} using broader supervision (1-4$s$ in the future). Meanwhile, it achieves further improvement when using a broader supervision range.}\label{tab:abla_dist}
\vspace{-5pt}
\end{table}

\subsubsection{Improvement with Broader Supervision Range}\label{sec:abla_better_sup}
As single-view reconstruction is supervised by multiple posed images, the performance can be improved by expanding the supervisory range during training. 
To this end, we investigate if our method can achieve consistent improvement using a broader supervisory range.
In standard supervision, we use fisheye views at the next time step (1$s$ in the future). In the broader supervisory range, we randomly incorporate fisheye views within the next [1-4$s$] timestep, yielding diverse supervisory ranges with a maximum coverage of 40m.
We compare with BTS \cite{wimbauer2023behind} in Tab.~\ref{tab:abla_dist}. Our method with the standard supervisory range produces comparable results to~\cite{wimbauer2023behind} with a broader supervision range. Enhanced by a broader range of supervision, our method achieves further improvement over~\cite{wimbauer2023behind}.

\subsection{Zero-shot Generalization on DDAD}\label{sec:exp_generalization}
We evaluate the zero-shot generalization of KITTI-360 trained models on the DDAD dataset. We report the $[0,50]$ meters range scene reconstruction scores in Tab.~\ref{tab:ddad}. Our method outperforms previous work, showing the effectiveness of the \ac{vl} guidance for zero-shot generalization.

\begin{table}
\centering
\footnotesize
\setlength{\tabcolsep}{5.5pt}
\begin{tabular}{ p{2.5cm}<{\raggedright} c c c}
\toprule[1.0pt]
{Method}  & O${_\text{acc}}$ & IE$_{\text{acc}}$ & IE$_{\text{rec}}$ \\
\midrule
PixelNeRF \cite{yu2021pixelnerf} & \nd 0.55 & \rd 0.45 & \nd 0.23 \\
BTS \cite{wimbauer2023behind} & \rd{0.48} & \nd {0.50} & \rd{0.16} \\
\textbf{Ours} & \fs{0.59} & \fs{0.58} & \fs{0.42} \\
\bottomrule[1.0pt]
\end{tabular}
% \vspace{-5pt}
\caption{\textbf{Generalization to DDAD with KITTI-360 trained model}. We evaluate in the [4, 50m] range. Our method demonstrates better zero-shot ability compared to previous methods.}\label{tab:ddad} 
\vspace{-10pt}
\end{table}

\section{Conclusion}\label{sec:conclusion}
In this paper, we proposed \ac{kyn}, a new method for single-view reconstruction that estimates the density of a 3D point by reasoning about its neighboring semantic and spatial context. To this end, we incorporate a \ac{vl} modulation module to enrich 3D point representations with fine-grained semantic information. We further propose a \ac{vl} spatial attention mechanism that makes the per-point density predictions aware of the 3D semantic context. Our approach overcomes the limitations of prior art~\cite{wimbauer2023behind}, which treated the density prediction of each point independently from neighboring points and lacked explicit semantic modeling. Extensive experiments demonstrate that endowing \ac{kyn} with semantic and contextual knowledge improves both scene and object-level reconstruction. Moreover, we find that \ac{kyn} better generalizes out-of-domain, thanks to the proposed modulation of point representations with strong vision-language features. The incorporation of \ac{vl} features not only enhances the performance of \ac{kyn}, but also holds the potential to pave the way towards more general and open-vocabulary 3D scene reconstruction and segmentation techniques.

{
    \small
    \bibliographystyle{ieeenat_fullname}
    \bibliography{main}
}

% WARNING: do not forget to delete the supplementary pages from your submission 
% \input{sec/X_suppl}

\end{document}